# Clever Search: A WordNet Based Wrapper for Internet Search Engines

Peter M. Kruse, André Naujoks, Dietmar Rösner, Manuela Kunze

Otto-von-Guericke-Universität Magdeburg, Institut für Wissens- und Sprachverarbeitung, P.O. Box 4120, D-39016 Magdeburg, Germany

roesner|makunze@iws.cs.uni-magdeburg.de

**Typ des Beitrags/Type of the paper**    Workshop

# Clever Search: A WordNet Based Wrapper for Internet Search Engines

Peter M. Kruse, André Naujoks, Dietmar Rösner, Manuela Kunze

This paper presents an approach to enhance search engines with information about word senses available in WordNet. The approach exploits information about the conceptual relations within the lexical-semantic net. In the wrapper for search engines presented, WordNet information is used to specify a user's request or to classify the results of a publicly available web search engine, like Google, Yahoo, etc.

In diesem Beitrag wird ein Ansatz vorgestellt, der auf der Grundlage der verfügbaren Informationen in WordNet die Ergebnisse von herkömmlichen Suchmaschinen verbessert. Es werden hierzu die konzeptuellen Relationen des lexikalischen-semantischen Netzes genutzt. Der beschriebene Suchmaschinen-aufsatz nutzt WordNet-Informationen um Nutzeranfragen zu spezifizieren und um die gefundenen Webseiten der herkömmlichen Suchmaschinen (Google, Yahoo etc.) zu klassifizieren und zu gruppieren.

## 1. Introduction

In most cases, when a user employs a web search engine, he will be confronted with a large amount of web pages as results. Most of the web search engines rank the web pages according to their relevance.

The vivísimo (http://vivisimo.com/) search engine offers the user web pages classified according to frequent words in the web pages. For example, as result for the search term 'Java', the user gets a list of web pages which are grouped into categories like *Java*, *Technology*, *JavaScript* etc.

But web pages about the topic 'Java' in the sense of 'island' or 'coffee' don't occur in the list. Other search engines deliver only a small number of web pages about the topics 'coffee' or 'island'. In this case, it is necessary to extend the search request by additional information.

In general, two deficiencies can be observed when using a common web search engine:

- web pages without the relevant information are presented in the results and

- web pages are not grouped according to similar content (classification).

To avoid these deficiencies, additional information for the query and a posteriori analyses of query results are necessary.

For the *first problem* described above, the extension of the user request is helpful. But the question is: Which additional terms added to the user request are necessary to get only relevant web pages?

The *second problem* is related to a user-friendly presentation of the results. In this case, it is sufficient to analyse the occurrence of relevant terms on the web pages within the result set of the query. Here again, it must be decided what the relevant terms on a web page are (with respect to the user request).

Both (problem) cases need information about the relevant terms: in the first case to expand the user query and in the second case for the classification of web pages within the answer set.

For the selection of relevant terms, the wrapper described below uses WordNet's information about different senses of a word. WordNet (Miller, G. (1990)) contains one or more senses for a word. For each sense there exists information about conceptual relations (like hypernyms, hyponyms, etc.). In this lexical-semantic net, each concept presented in a conceptual relation is represented by a so-called synset. A synset is a set of synonyms, which can contain more than one element. The wrapper uses these words to improve the results from a common web search engine and their presentation to the user.

The next section describes the wrapper and its different modes. After this, some ideas are outlined for an improvement of the post-filter mode of the wrapper. In section 4, we describe the integration of GermaNet (Hamp, B., Feldweg, H. (1997), Kunze, C. (2001)) as resource for the wrapper. A summary is given at the end of the paper.

## 2. Clever Search

The wrapper for web search engines described in this paper supports the user in two modes in order to cope with the problems described above: *search with a pre-filter* and *search with a post-filter*. Both approaches use information available from WordNet in order to improve the results of a standard web search engine.

The wrapper can be used with different common search engines (Google, Yahoo, and MSN). For search engines that use the same sources (e.g. Yahoo and Altavista), only one search engine of this group is offered to the user. Via configuration parameters, additional search engines can also be integrated into the *clever search* wrapper.

In the following, both modes of the wrapper are explained in detail.

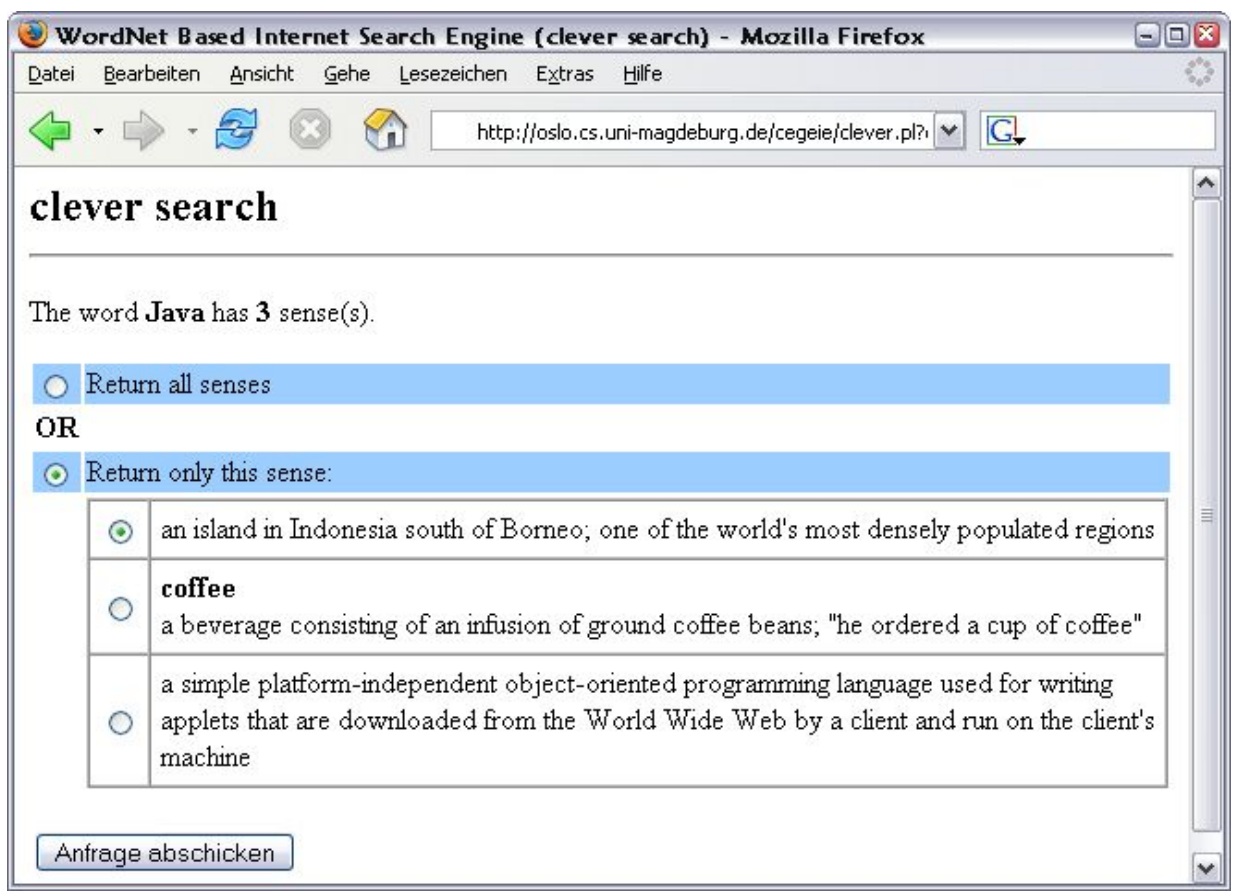

Fig. 1: Selection of word senses for a request.

## 2.1. Search with a pre-filter

When searching with a pre-filter, the user request will be extended by additional terms to obtain better search results. The user can concretise the search request by selecting a distinct word sense from WordNet (see Fig. 1). For the selection process, WordNet's short glossary descriptions for each sense are presented to the user. The user has two options: the user can either use the option 'return all senses' or only one specific sense can be selected.

If the option 'return all senses' is selected then no extension is carried out. In this case, the original user request will be forwarded to the search engine.

In the other case, when one specific sense is selected, the original user request is extended by information selected from WordNet. This request will be forwarded to the search engine and the results of the search are presented by the search engine.

For example, the user types 'ring' as input. *Clever search* presents all senses, like ring → gang; ring → jewelry; etc. The user selects one sense and the user's original request will be extended with the WordNet information of the selected sense (e.g. all pages about ring → jewelry).

Another example: in the example interaction (cf. Fig. 1), three different WordNet senses of the query term 'Java' are offered. For each choice of a specific sense, the user gets a different cluster of result pages (cf. Fig. 6 to Fig. 8).

For the extension of the user request, WordNet information about terms in conceptual relations with the selected sense is inserted into the request. First the hyponymy relation is checked and the words which are members of the related synsets are used. If no such information about hyponyms exists then the hypernymy relation is exploited for the extension.

## 2.2 Search with a post-filter

In this case, the original user request will be sent to a search engine without any changes, but the results of the search will be classified according to information available from WordNet. The user can choose whether all word senses should be used for this analysis or only one sense (Fig. 2). If the user selects only one sense then the web pages found by a standard search engine are analysed with respect to the selected sense. Otherwise, the calculation of the weights are carried out for each sense available in WordNet.

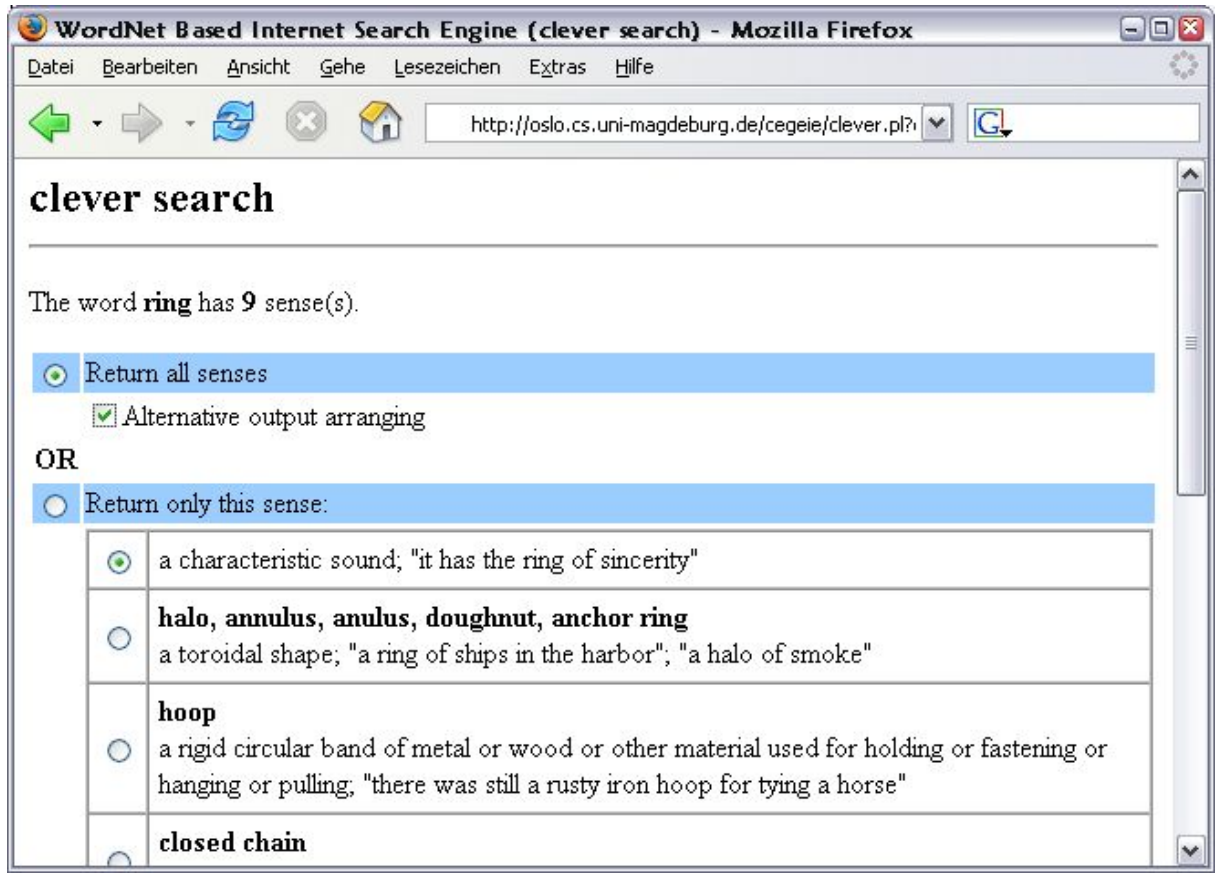

Fig. 2: Selecting form for post-filter mode.

The classification process for retrieved documents is based on weighting the number of occurrences of concepts from WordNet related to the input. All information available via hypernymy, hyponymy, meronymy, and holonymy relations for the different senses is used as relevant concepts.

The results of this mode can be presented in list form or in a ranking table. The screenshot presented in Fig. 3 shows a ranking table for the search string 'ring'. In WordNet, 9 senses are available for this term. For each web page found by a standard search engine, the weight was calculated with respect to the information for the different senses within WordNet (because the option 'Return all senses' was selected). In the ranking table, all web pages and their weights for the different senses (columns '1.' to '9.') are presented. Above the table, the short glossary descriptions of the different senses are given.

The weighting process itself will be explained below.

*Weighting of Results for Classification*

The next paragraphs describe the steps of the weighting algorithm, which are realized for each WordNet sense. For each sense of a search term in case of first option one or for the chosen sense (in case of second option), a list is created of all concepts of hypernyms, hyponyms, meronyms, and holonyms of a sense. Then according to the position in the tree of conceptual relations, the concepts are weighted.

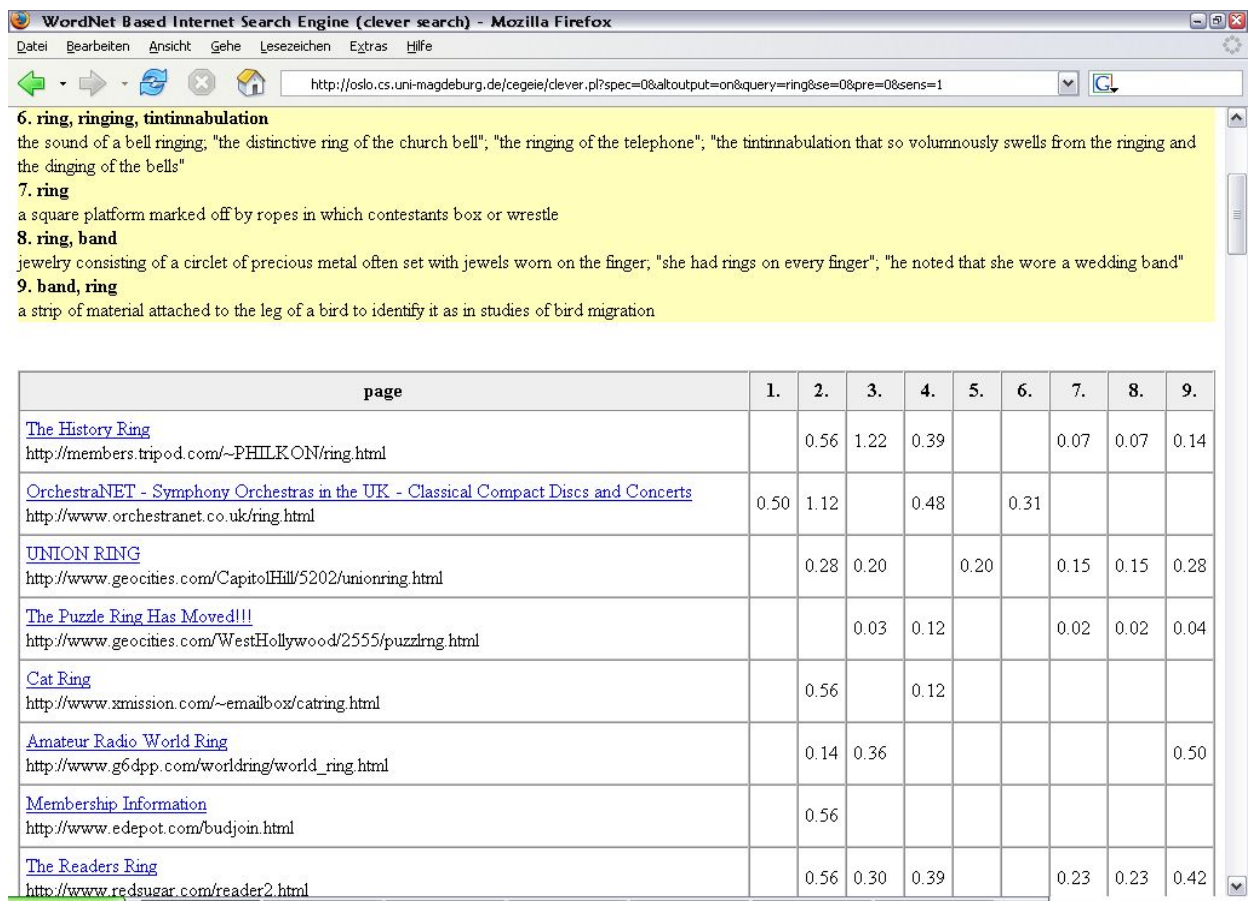

Fig. 3: Ranking table of the results.

In the next step, each web page delivered by the common search engine will be analysed with respect to each of the possible word senses. The occurrences of terms referring to different concepts on the web page are counted. After this, the number of occurrences is multiplied by the weight of the concept. The weight of the web page is the sum of the weight of all concepts appearing within the web page.

For each sense, the web pages can be listed in descending order according to their weight for this sense.

A crucial point in the weighting process is the determination of the relevance factor. Using the level information within the hypernym tree is only a coarse approach. This can be improved by using and applying more information for the determination of the relevance factor to get a more differentiated factor. In the next section, a number of ideas for this are presented.

## 3. Improvement of the Algorithm

There are a number of ways to improve the algorithm. The following list contains some ideas which can be integrated into the algorithm:

- Even for different senses there may exist joint ancestors in the hypernym tree. In this case, the terms associated with these (general) concepts should get a lower relevance factor than those of a more specific concept (occurs only in one sense of WordNet). For example, the concept 'entity' of the hypernym tree of the entry 'Java' (see Fig. 4) gets a lower relevance factor than the concept 'abstraction', because the former is defined as hypernym for two distinct senses ('island' and 'coffee').

- Another extension of the algorithm could be a user-definable weighting of the different conceptual relation types. For example, hypernyms or hyponyms could be assigned a higher relevance factor than meronyms.

- Short web pages often contain information in a more compact and more condensed form. This means, the information on these web pages is probably more relevant for a given topic. On the other hand, for a web page with a longer text it can be assumed that this web page is less focussed and that it can also contain concepts which are not so relevant for a topic.

- If one sense is selected, then it is possible to use the information about all senses. The concepts of the non-selected senses are assigned with a negative factor, i.e., if one of these concepts occur in the web page then the negative factor is included in the calculation of the weight of a web page. Thus web pages with content about another (non-selected) sense or web pages with multi-thematic content get a lower weighting, while web pages with content about the sense selected obtain a higher weighting.

- Another finer classification approach is to consider the word frequencies within a (sub-)language. Frequent words are assigned with a lower relevance factor than rarely used words. Statistical approaches proven in many information retrieval (IR) applications can be used for this task. For example, the integration of IR measures (Sparck Jones (1972)) like term frequencies (TF) and inverse document frequencies (IDF) will be helpful for the weighting process within the *Clever Search* wrapper.

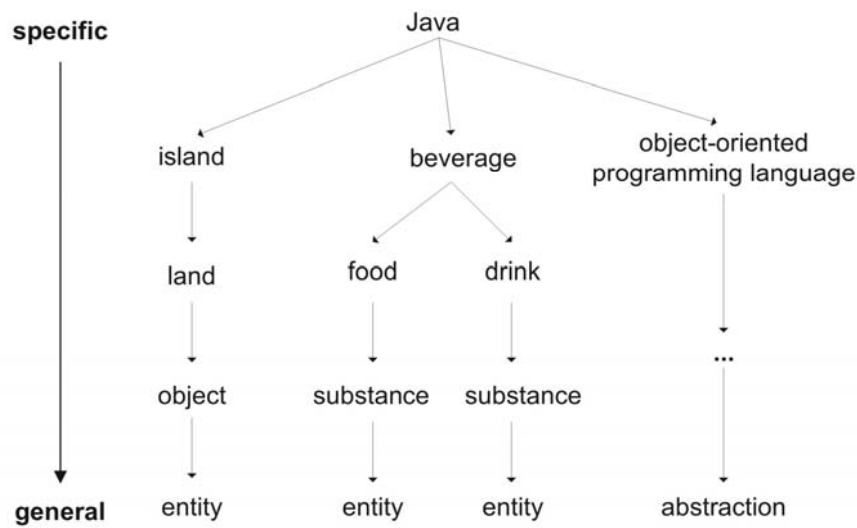

Fig. 4: Excerpt of the hypernym tree for the concept 'Java'.

## 4. Integration of GermaNet

Can *Clever Search* be used with GermaNet for retrieving web pages in German? In both methods (pre-filter and post-filter), the WordNet function 'Overview of Senses' (option '-over') and the parameter –g ('Display gloss') is used. This functionality delivers information about the different senses for the different word categories (noun, verb, etc.). For each sense, a short natural language description is presented (see Fig. 5).

This function is a feature of WordNet version 2.0. The current version of GermaNet does not support such a function. Additionally, the short natural language descriptions are often missing in GermaNet. Therefore an easy, direct adaptation of the wrapper for GermaNet is not possible.

Nevertheless, it is possible to use the wrapper with GermaNet. One solution could be that instead of the description of the word sense, information about the hypernymy and hyponymy relation is presented to the user, so that the user can choose one of the senses. Further more, the information presented by the 'overview' function in WordNet can be created manually. It is a little bit more complex than to use a predefined function, but it is possible.

## 5. Summary

The wrapper for search engines presented in this paper uses WordNet in two modes to improve the results. One mode expands the user's request with additional information based on WordNet's senses. The other mode classifies the results of a search engine according to WordNet's sense information. Both approaches exploit WordNet's information about the conceptual relations for a word sense.

The search with a pre-filter delivers better results than the search with a post-filter. The reason is that in general it is not possible to analyse **all** results of the search engine. In many cases, the search engine delivers only its 1000 top-ranked pages. The search with a post-filter will possibly improve when all pages found can be analysed.

The further development of the wrapper will include the integration and evaluation of different weighting strategies as described in section 3. Another point which should be investigated is the combination of both modes, so that the user can select a sense and the web pages delivered by the standard search engine will be ranked with respect to the information available in WordNet.

The *Clever Search* wrapper described in this paper is available at http://oslo.cs.uni-magdeburg.de/~cegeie/

The techniques from Clever Search's pre-filter approach are useful for the retrieval of other media types as well. Currently, only text retrieval is implemented, but experiments with query expansion for image retrieval in pre-filter mode showed promising results (cf. Fig. 9).

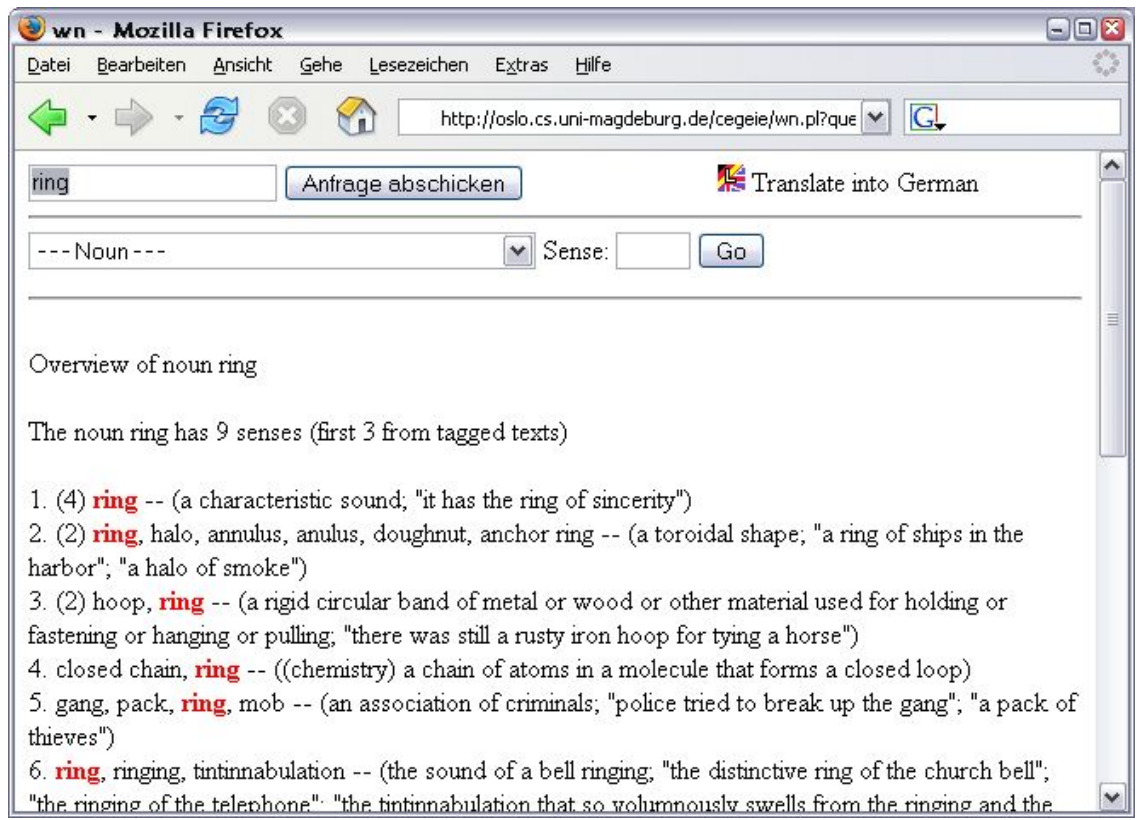

Fig. 5: Function 'overview' for the term 'ring'.

## Literaturverzeichnis/References

Miller, G. (1990). Five Papers on WordNet. CSL-Report (43). Cognitive Science Laboratory. Princeton University.

Hamp, B., Feldweg, H. (1997). GermaNet - a lexical-semantic Net for German. In Vossen, P. et.al.: Proc. of ACL/EACL-97 workshop Automatic Information Extraction and Building of Lexical Semantic Resources for NLP Applications. 9-15, Madrid.

Kunze, C. (2001) Lexikalisch-semantische Wortnetze, 386-393, Spektrum Akademischer Verlag, Heidelberg; Berlin.

Sparck Jones, K. (1972). A Statistical Interpretation of Term Specificity and its Application in Retrieval. Journal of Documentation, 28:1. 11-21.

# Appendix

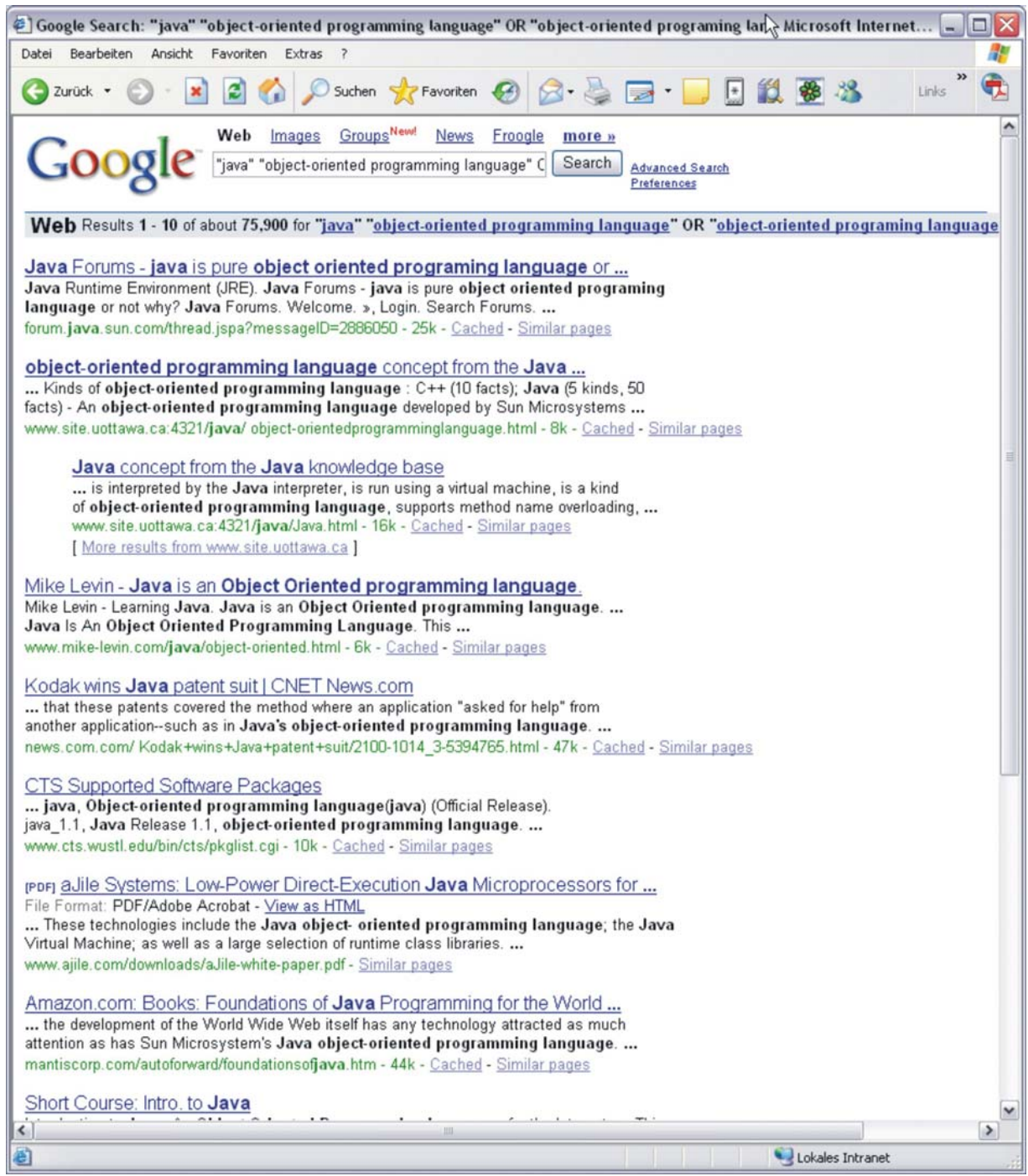

Fig. 6: Results of Clever Search for WordNet's sense of 'java': 'programming language'.

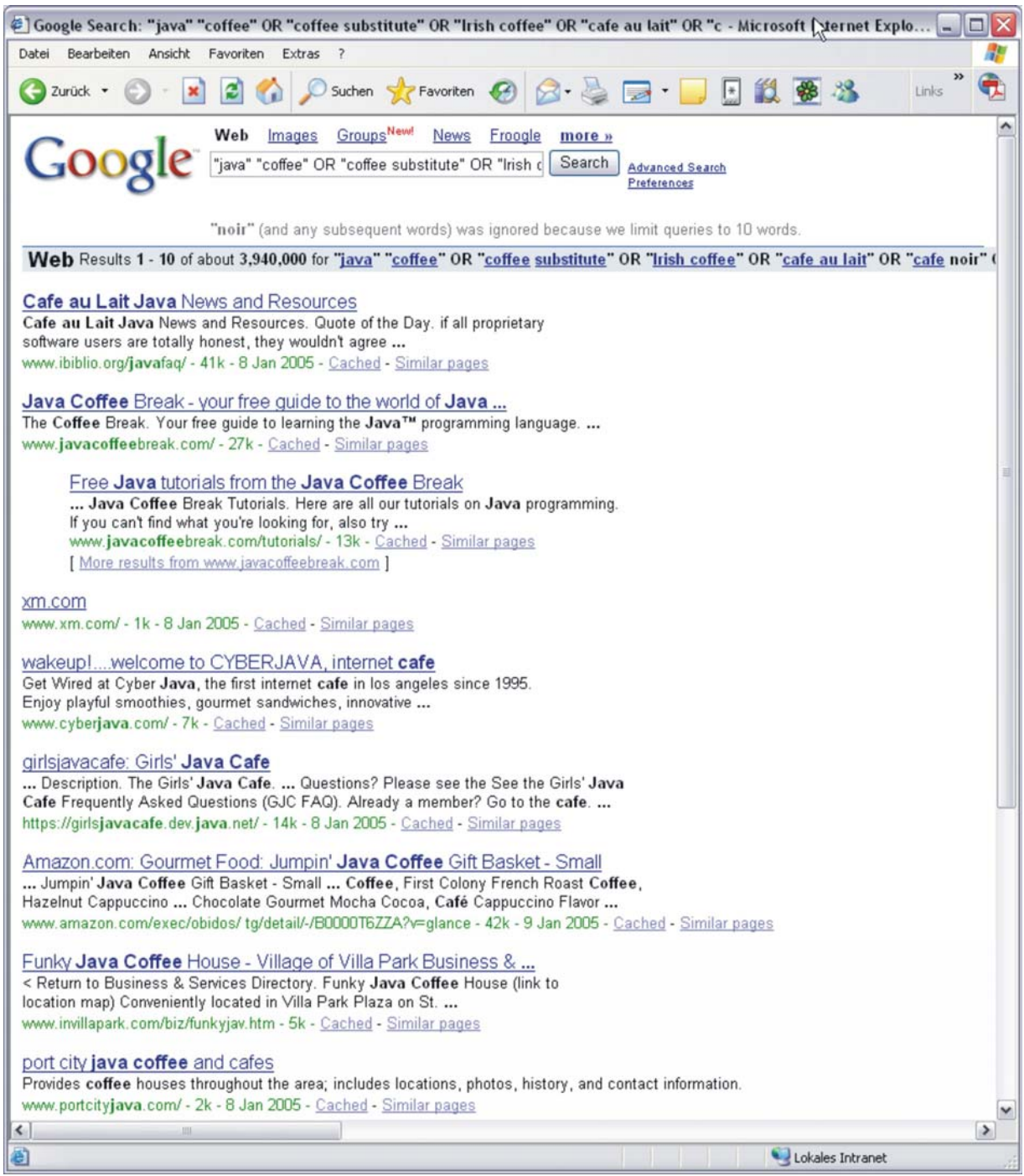

Fig. 7: Results of Clever Search for WordNet's sense of 'java': 'coffee'.

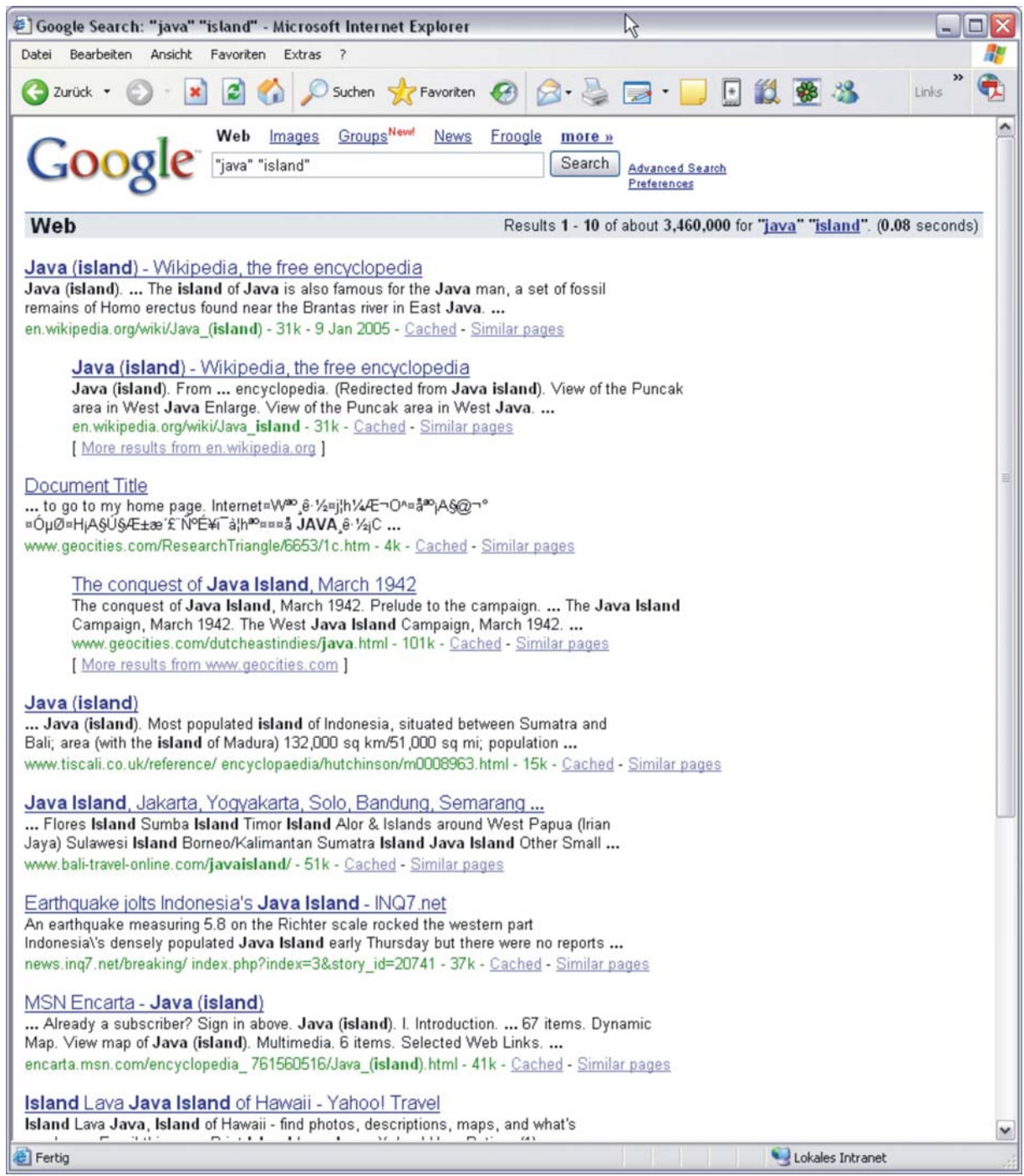

Fig. 8: Results of Clever Search for WordNet's sense of 'java': 'island'.

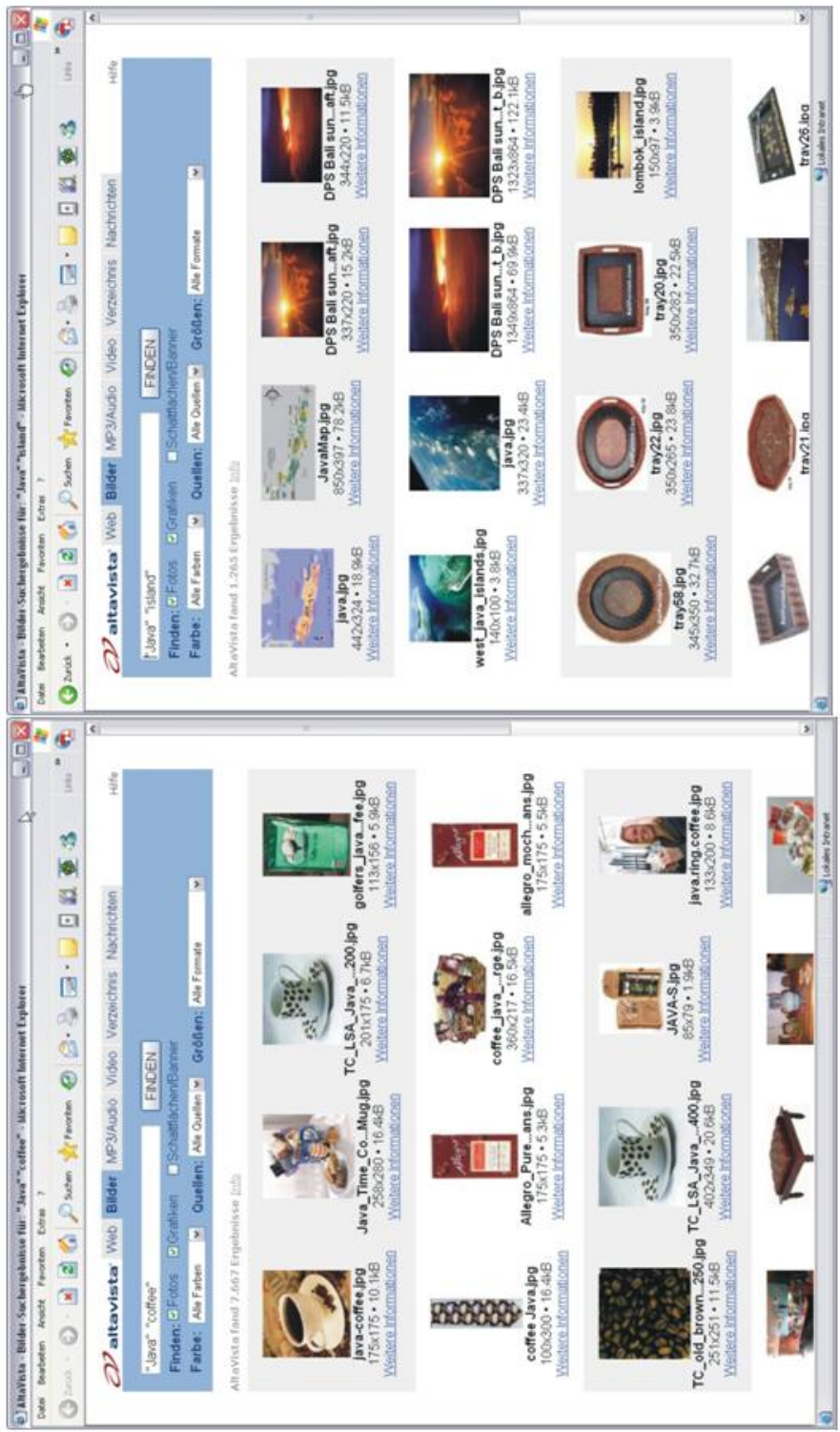

Fig. 9: Image retrieval for 'java' for senses 'coffee' and 'island'.